\pdfoutput=1
\documentclass[11pt]{article}

\usepackage[]{EMNLP2023}

\usepackage{times}
\usepackage{latexsym}

\usepackage[T1]{fontenc}
\usepackage[utf8]{inputenc}

\usepackage{microtype}

\usepackage{inconsolata}
\usepackage{algorithm}
\usepackage{algorithmic}
\usepackage{multirow}
\usepackage{tabularx}
\usepackage{booktabs}

\usepackage{graphicx}
\usepackage{amssymb}
\usepackage{amsmath}

\title{PlugMed: Improving Specificity in Patient-Centered Medical Dialogue Generation using In-Context Learning}

\author{Chengfeng Dou ~Zhi Jin\thanks{*Corresponding authors.} ~Wenpin Jiao\thanks{*Corresponding authors.} ~Haiyan Zhao \\ 
~{\bf Yongqiang Zhao} ~{\bf Zhenwei Tao}\\
        School of Computer Science, Peking University;\\ 
        Key Laboratory of High Confidence Software Technologies(PKU), MOE of China \\
        \texttt{\{chengfengdou,zhijin,jwp,zhhy.sei\}@pku.edu.cn}\\
        ~\texttt{\{tttzw,yongqiangzhao\}@stu.pku.edu.cn}}

\begin{document}
\maketitle
\begin{abstract}
% The patient-centered medical dialogue systems strive to offer diagnostic interpretation services to users who are less knowledgeable about medical knowledge, emphasizing the importance of providing responses specific to the patients.
The patient-centered medical dialogue systems strive to offer diagnostic interpretation services to users who are less knowledgeable about medical knowledge, through emphasizing the importance of providing responses specific to the patients.
% Although large language models (LLMs) have promising performance even in some tasks in medical field, it is difficult to guarantee the specificity of responses.
It is difficult for the large language models (LLMs) to guarantee the specificity of responses in spite of its promising performance even in some tasks in medical field.
Inspired by in-context learning, we propose PlugMed, a Plug-and-Play Medical Dialogue System, for addressing the challenge.
% PlugMed enhances LLMs' dialogue strategies with two modules, the prompt generation (PG) module and the response ranking (RR) module, to ensure the specificity of the dialogue.
PlugMed is equipped with a prompt generation (PG) module and a response ranking (RR) module to enhances LLMs' dialogue strategies for improving the specificity of the responses.
%To address this challenge, we draw inspiration from in-context learning and propose PlugMed, a Plug-and-Play Medical Dialogue System that enhances appropriate dialogue strategies of LLMs through two modules: the prompt generation (PG) module and the response ranking (RR) module.
The PG module is used to stimulate the imitative ability of LLMs by providing them with real dialogues from similar patients as prompts. 
% The RR module incorporates fine-tuned SLMs as response filters, enabling the selection of appropriate responses generated by LLMs. 
% The RR module incorporates fine-tuned small model as response filter, enabling the selection of appropriate responses generated by LLMs. 
The RR module incorporates fine-tuned small model as response filter to enable the selection of appropriate responses generated by LLMs.
% Furthermore, we introduce a new evaluation method based on intent and high-frequency medical term matching to effectively assess the specificity of the responses.
Furthermore, we introduce a new evaluation method based on matching both user's intent and high-frequency medical term to effectively assess the specificity of the responses.
% Experimental evaluations conducted on three medical dialogue datasets, including both automatic and manual assessments, demonstrate the effectiveness of our approach.
We conduct experimental evaluations on three medical dialogue datasets, and the results, including both automatic and human evaluation, demonstrate the effectiveness of our approach.
\end{abstract}
\section{Introduction}
%Patient-centered medical dialogue generation~(PMDG) holds significant potential for reducing diagnostic costs and enhancing medical efficiency.
% Medical Dialogue Generation aims to automatically generate informative responses which is a key task in health conversational assistants. 
As a key task in health conversational assistants, Medical Dialogue Generation aims to automatically generate informative responses to the users.
% Such dialogues have to be both medical knowledgeable and patient-specific, as the patients may be less knowledgeable about medical knowledge and their dialogue contexts may vary.
It's better for such dialogues to be medical knowledgeable, patient-specific and context-aware, as the patients may be less knowledgeable about medical knowledge and their dialogue contexts may vary.
%and their dialogue contexts vary frequently.
%It is designed to provide specific diagnostic feedback to users who may have little medical knowledge.
%Prior studies~\cite{Li2021, varshney2023a, varshney2023b} have emphasized the critical role of domain knowledge in medical dialogue generation. 
% any of these studies utilize knowledge injection modules to integrate general medical knowledge from knowledge bases into the models. 

Prior studies~\cite{Li2021, varshney2023a, varshney2023b} have emphasized the critical role of domain knowledge in medical dialogue generation, many of which integrate general medical knowledge from knowledge bases into the models by utilizing knowledge injection mechanism.
% With the growing size of models, the problem of knowledge scarcity is being mitigated. 
%However, the lack of knowledge is being alleviated as large language models~(LLMs) continue to grow in size, implying that the focus of medical conversation research may be changing.
%This is evident in the performance of LLMs such as Instruct GPT~\cite{Long2022}, which outperform small fine-tuning-based models on medical Q\&A tasks without any training~\cite{singhal2022large}. 
%This demonstrates the great potential of LLMs in medical dialogue.
As the scale of large language models~(LLMs) continues to expand, the lack of knowledge is being alleviated. 
This is evident in the performance of LLMs such as Instruct GPT~\cite{Long2022}, which outperform small models based on fine-tuning ~\cite{singhal2022large} on medical question answering tasks without any additional training.

\begin{figure}[t]
\centering
\includegraphics[width=0.48\textwidth]{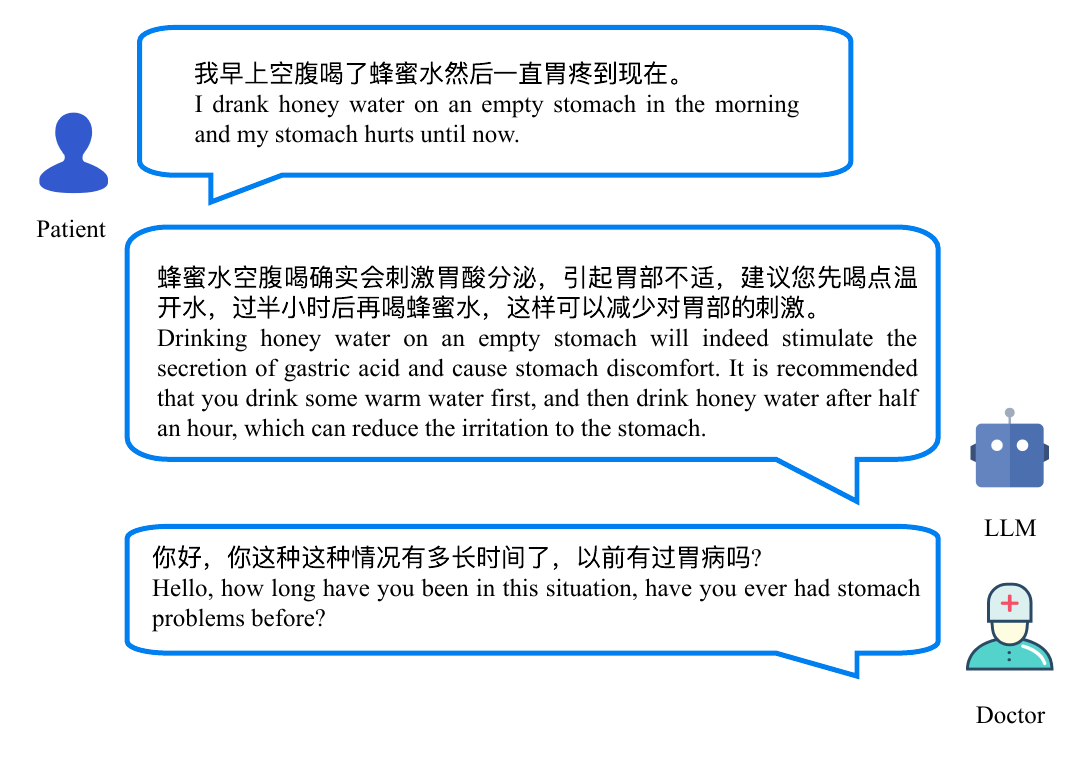} 
\caption{An example of the medical dialogue. The LLM and the doctor adopt different response strategies for the same question.}
\label{fig:cover}
\end{figure}

% Despite their extensive domain knowledge, LLMs have been criticized for lacking medical diagnostic logic to the patient they are facing, as this can lead to untargeted and even risky response suggestions~\cite{HOWARD2023405, huatuogpt-2023}. 
However, LLMs have been criticized for lacking medical diagnostic logic to the patient, which can lead to untargeted and even risky response suggestions~\cite{HOWARD2023405, huatuogpt-2023}.
% For instance, as illustrated in Figure~\ref{fig:cover}, when a patient claims to have a particular ailment and requests medication, the doctor's priority is to first investigate the underlying disease in order to make targeted suggestions.
For instance, as illustrated in Figure~(\ref{fig:cover}), when a patient claims to have a particular ailment and seeks medication, the doctor's priority is to first investigate the underlying disease to make targeted suggestions.
%rather than providing immediate medical advice. 
In contrast, LLMs are more inclined to give straightforward medical advice, rather than further gathering patient information to give accurate advice.
%In contrast, LLMs tend to provide direct, generalized medical advice rather than gathering further necessary information specific to the patient. 
% This generalized and potentially ambiguous information places a burden of understanding on the patient.
%This means that patients must know their condition well in order to use LLMs effectively, which is contradictory to their lack of medical knowledge.
% Such overconfident behaviours of LLMs places a lot of understanding burden on the patient.
Once LLMs give overconfident advice, it is difficult for patients to discern the effectiveness and safety of such advice.

% 我这里直接讲微调的部分去掉了，只强调 ICL 能够做到类似的事情。
% Previous research has demonstrated that in-context learning~(ICL) possesses the ability to impact the conversational style of language models (LLMs) and mitigate prejudice and toxicity concerns~\cite{Roy2023,Meade2023} by giving a few suitable examples. 
Researches have demonstrated that in-context learning~(ICL) possesses the ability to impact the LLMs' conversational style and mitigate prejudice and toxicity concerns~\cite{Roy2023,Meade2023} by demonstrating a few examples. 
This phenomenon is due to the powerful imitation learning and few-shot capabilities of LLMs, by learning patterns from a small number of samples and applying them in generation~\cite{Dong2022}.

% Taking inspiration from these studies, we propose leveraging ICL to shape the LLMs' dialogue strategies and accordingly we introduce a Plug-and-Play Medical Dialogue System, PlugMed, comprising two crucial components: a prompt generation (PG) module and a response ranking (RR) module.
Taking inspiration from these studies, we propose leveraging ICL to shape the LLMs' dialogue strategies and accordingly design a Plug-and-Play Medical Dialogue System, named PlugMed, which embodies two crucial components: a Prompt Generation~(PG) Module and a Response Ranking~(RR) Module.
Specifically, PlugMed uses the PG module to identify examples by considering information from both global and local views. 
%$ When considering the global view, we exploit the similarity of the entire dialogue history to choose relevant examples. 
% The global view exploits the similarity with the entire dialogue history to choose relevant examples, for ensuring that the model acquires a comprehensive understanding of the entire dialogue process. 
From global view, the PG module choose relevant examples for ensuring that the model acquires a comprehensive understanding of the entire dialogue process by exploiting the similarity with the entire dialogue history. 
% Conversely, when examining the local view, our design prioritizes recent utterances to capture the most relevant information for generating responses. 
Conversely, from local view, the PG module priorities recent utterances to capture the most relevant information for generating responses.
% While, the RR module utilizes a fine-tuned small model to autonomously select the most appropriate response for the ongoing dialogue, in order to further maximize the advantages of both the global and local perspectives, 
To further maximize advantages of both the global and local views, PlugMed uses the RR module to autonomously select the most appropriate response for the ongoing dialogue through utilizing a fine-tuned small model.

% When construct PlugMed, another critical consideration is the introduction of appropriate automatic evaluation metrics. 
% Another critical consideration is about appropriate automatic evaluation metrics when construct medical dialogue systems. 
Another critical consideration is about appropriate automatic evaluation metrics for medical dialogue systems. 
Previous studies~\cite{varshney2023a, huatuogpt-2023} only relied on open-domain dialogue evaluation methods.
However, as indicated in ~\cite{Ji2022}, these evaluation methods may be unreliable in task-oriented scenarios. 
% To gain a comprehensive understanding of the system's real-world performance, we undertake a thorough evaluation encompassing two key aspects: the intent accuracy and the high-frequency medical term accuracy. 
To gain a comprehensive understanding of the system's real-world performance, we undertake a thorough evaluation that is twofold: the intent accuracy and the high-frequency medical term accuracy. 
Here, the intent accuracy is used to evaluate the reasonableness of the dialogue actions adopted by the system, and the high-frequency medical term accuracy focuses on measuring the presence of essential medical information in the system's responses.

% We evaluate our approach on three datasets, i.e., Meddg, MedDialogue and Kamed datasets.
We evaluate our approach on three widely used large medical datasets, i.e., Meddg, MedDialogue and Kamed datasets.
% Both automated and manual evaluations show that our approach are effective.
Both automatic and human evaluations show that our approach can substantially improve the specificity of LLMs.
Our contributions can be summarized as follows:
\begin{itemize}
    % \item An ICL-based approach enables LLMs to generate responses that conform to the diagnostic strategy.
    % \item Comprehensive evaluation metrics for medical dialogue automation that can consider both intent accuracy and high-frequency medical term accuracy.
    % \item An thorough experiment to demonstrate key elements of automatic medical diagnosis.

    \item An ICL-based approach that enhances LLMs to generate responses that conform to the diagnostic strategy.
    \item The comprehensive evaluation metrics for medical dialogue automation that take both the intent accuracy and the high-frequency medical term accuracy into account.
    \item Experiments demonstrating the key elements of automated medical diagnosis.
\end{itemize}
\section{Methodology}
\begin{figure*}[!ht]
\centering
\includegraphics[width=0.99\textwidth]{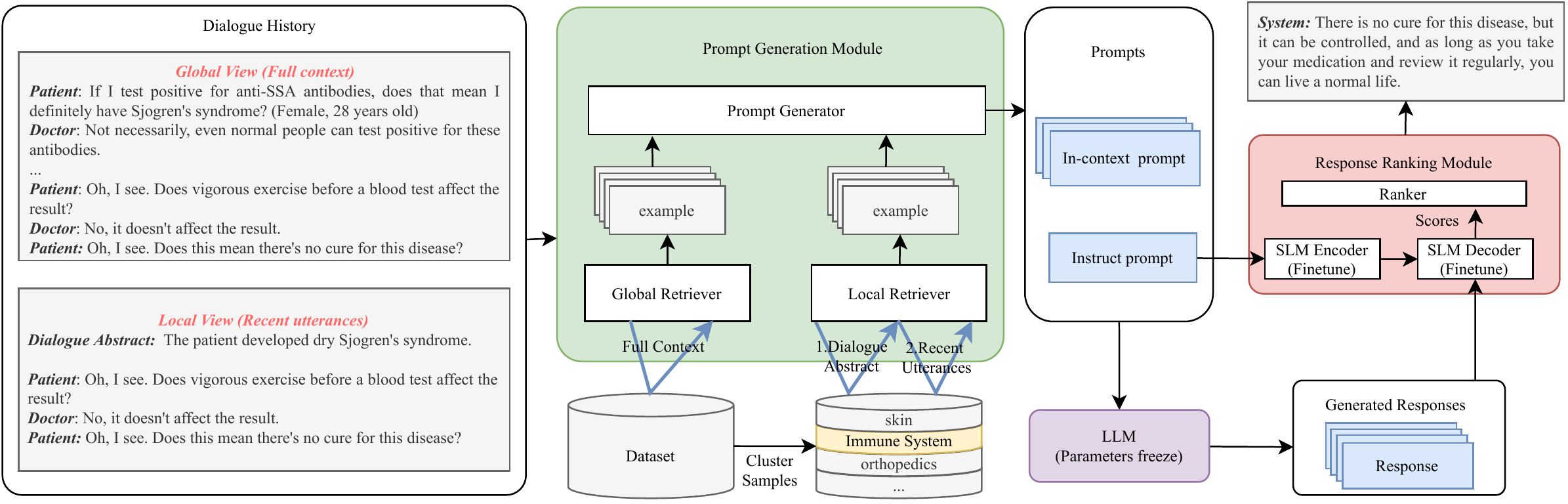} 
% 修改描述
% \caption{The overview of PlugMed. Our system consists a Prompt Generation module and a Response Ranking module.}
\caption{
The overview of PlugMed. 
Our system consists of two core components, i.e., \textit{Prompt Generation (PG) Module}, which retrieves similar examples in the dialogue history from both global and local views to generate prompts, and \textit{Response Ranking (RR) Module}, which ranks the outputs of LLM corresponding to these prompts and selects the best responses.}
% 这个图里面紫色的明显也是一个module，而且不可或缺，虽然你核心设计了两个模块，但加上紫色的才是一个完整的流程。这个地方紫色的要不要说明一下？现在这种说法感觉是不完整。

\label{fig:structure}
\end{figure*}

\subsection{Overview}
%To enhance the dialogue strategy of a LLM, 
% We propose a scheme that employs ICL to guide the LLMs towards generating high-quality replies by enhancing the dialogue strategy and utilizes a ranking model to filter the model's responses. 
%Additionally, we utilize a ranking model to filter the model's responses. 
%An overview of our approach, illustrated in Figure~\ref{fig:structure}, comprises the Prompt Generation (PG) Module and the Response Ranking (RR) Module.
We propose a framework as shown in Figure~(\ref{fig:structure}), which employs ICL to guide the LLM towards generating high-quality replies, where the Prompt Generation (PG) Module and the Response Ranking (RR) Module are two key components.
% As illustrated in Figure~\ref{fig:structure}, two key components in our scheme are the Prompt Generation (PG) Module and the Response Ranking (RR) Module.
The PG Module accepts the dialogue history as input and outputs multiple In-context prompts, along with a single Instruct prompt. 
%Then, these prompts are fed into the LLM, and 
%These prompts are fed into the LLM to generate multiple system responses.
Based on these prompts, the LLM generate multiple system responses.
% Finally, the RR Module is employed to identify the optimal response.
Then, the RR Module utilizes a small language model~(SLM) to select the best response.
% In the following subsections, we will detail these two components.
We will elaborate these two components in the following subsections.

\subsection{Prompt Generation Module}
The PG module uses a multi-strategy retrieval framework to retrieve examples similar to the input sample and then employ them to generate prompts.

\subsubsection{Basic Ideas} \label{sec:bi}
% When generate prompts, it is necessary to retrieve relevant examples similar to the input dialogue sample. 
%To leverage the dialogue information more effectively, the retrieval framework capture the information from both global and local views. 
As shown in `Dialogue History' at the left part of Figure~(\ref{fig:structure}), the main idea is as follows.
% Firstly, the global view considers the entire history of the dialogue, and the global retriever retrieves the similar dialogues from the raw dataset.
Firstly, from the global view, we considers the entire history of the dialogue, and the global retriever retrieves the similar dialogues from the the training set of the dataset.
%As depicted in Figure~(\ref{fig:structure}), in Dialogue History, the global view entails considering the entire history of the dialogue to retrieve similar dialogues from the original dataset. 
%This approach allows us to leverage all the information available in the conversation. 
Secondly, considering the global view is susceptible to distractions caused by the abundance of irrelevant information, %such as small talk, 
which may lead to inappropriate retrieved examples, 
we include the local view for enhancing the relevance of the retrieval.
%However, we find that the global view is susceptible to distractions caused by the abundance of irrelevant information, such as small talk, leading to inappropriate retrieved examples.
%To enhance the relevance of the search process, we propose another retrieval approach from a local view. 
Concretely, the local retriever first extracts the patient's symptom information from past conversations, serving as the initial filter for the samples. 
% It then concentrates on the most recent rounds of conversations, as these conversations hold the utmost relevance to the system-generated responses.
% We then utilize the recent rounds of conversations as query to select examples, as these conversations hold the utmost relevance to the system-generated responses.
Considering the recent rounds of conversations hold the utmost relevance to the system-generated responses, we then utilize these conversations as query to select examples.
This effectively mitigate the interference of irrelevant information.
Thirdly, 
%while the local view may have limited information due to its disregard for a significant portion of the conversation history, 
we employ the retrieved examples from both views to take advantage of their respective strengths.
% However, relying solely on the global perspective is not enough. 
% Lengthy dialogues sometimes contain a large amount of irrelevant information such as chitchats in current dialogue generation task may hinder the retrieval of relevant examples. 
%For this reason, we also introduce the local perspective, focusing on recent rounds of dialogue, and we will describe its details in the following chapters.
% The local view is included for tackling this issue by focusing on recent rounds of dialogue.

\subsubsection{Implementation}
% \paragraph{Example Retriever.}
\paragraph{Global Retriever.}
% The global retriever uses the full context of dialogue history as query to search samples. 
% To achieve this, we employ Sentence-Bert~\cite{sentence-bert} to independently encode the entire conversation histories of dialogue sample and examples. 
% And we use cosine similarity is utilized to identify the closest $k$ examples.
The global retriever utilizes the full context of dialogue history as a query for searching samples.
It employs Sentence-Bert (SBERT)~\cite{sentence-bert} to encode the query and examples, and then utilizes cosine similarity to identify the closest examples.
% However, the local view does not allow for a straightforward application of the same approach.
% That is because the last few rounds of dialogue provide insufficient information to unveil the patient's symptoms and disease details.
% To extract the relevant information from the early conversation history, we developed a medical dialogue summary model using the ICMS-MRG~\cite{icms} dataset with Bart~\cite{bart} as the skeleton model. 
% By leveraging this model, we are able to extract the chief complaint of the patient, effectively achieving our goal of compressing the conversation history while filtering out irrelevant information.
% Hence, to facilitate the retrieval of examples from a local view, we establish a two-level index retriever. 
% In contrast, the local view overlooks a substantial portion of conversation histories, making the direct retrieval method employed in the global view inadequate for ensuring the similarity of symptoms and disease information in the given examples. 
% Therefore, we propose an alternative retrieval approach consisting of two steps. 
% Firstly, the method retrieves a cluster of cases that display similar symptoms to those of the patient. 
% Secondly, We use Sentence-Bert to retrieve relevant examples from this set, with a specific emphasis on recent conversational rounds.

\paragraph{Local Retriever.}
The local retriever retrieves samples in terms of symptoms and recent utterances.
For getting symptoms, we develop a medical dialogue summary model that utilizes the ICMS-MRG~\cite{icms} dataset in conjunction with BART~\cite{bart} as the backbone model. 
This model enables the extraction of the chief complaint, containing the patient symptom information, and we use SBERT as the encoder of chief complaints to provide embeddings for the following operations.
We first encode the chief complaints of examples and use the K-Means algorithm to divide them into $\mathcal{K}$ groups.
Then, we extract the embeddings of each cluster centers to serve as the symptom index for querying.
When performing the search, we first retrieve the candidate examples by computing the cosine similarity between the embeddings of the sample's chief complaint and the symptom index.
% Then, we use SBERT to retrieve examples from the candidates base on recent dialogue utterances.
Then, we use SBERT to retrieve examples from the candidates based on recent dialogue utterances.

\paragraph{Prompt Generator.}
The prompt generator generates two types of prompts, as depicted in Figure~(\ref{fig:structure}): `Instruct prompt' and `In-context prompts'. 
% It is worth noting that we generate separate prompts for the global view and the local view.
% We ensure that examples obtained through different search strategies are not included in the same prompt.
% Hence, we generates multiple In-context prompts simultaneously.
Each In-context prompt corresponds to a distinct example retrieval strategy.

It is needed to compress examples to include more demonstrations, given the input length constraint imposed by the LLM, when generating the in-context prompts. 
Drawing inspiration from \citet{Hu2022}, for each example, we keep only the most recent rounds of conversations, and we restrict the maximum conversation length to no more than $n$. 
Moreover, we employ the previous mentioned chief complaints (up to $m$ characters long) as the dialogue abstract, replacing the excluded history to achieve dialogue compression. 

% The Instruct prompt includes the complete dialogue history, and we include the instruction before the history to prompt LLM to act as a doctor. 
The Instruct prompt includes the full context, and we include the instruction before the history to prompt LLM to act as a doctor. 
We use this prompt to activate the zero-shot capability of LLMs. 
Appendix~\ref{sec:prompt} gives some examples of both prompts.

\begin{figure}[ht]
\centering
\includegraphics[width=0.48\textwidth]{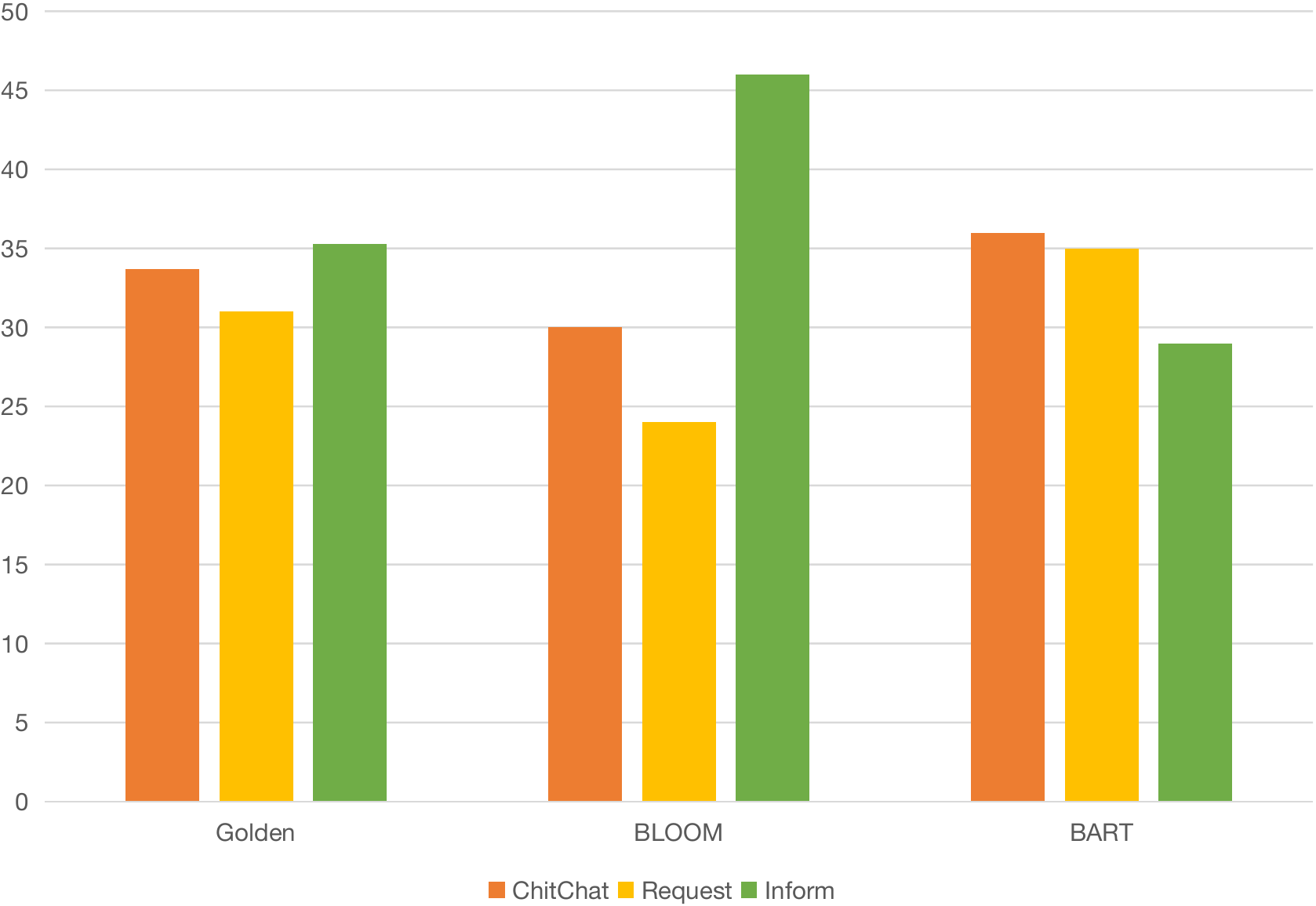} 
\caption{
We compared the behavior patterns of LLMs and SLMs, using Bloom~\cite{bloom} as a representative of LLMs and Bart~\cite{bart} as a representative of SLMs. 
We counted the distribution of dialogue actions taken by them on the validation set of KaMed~\cite{kamed}, where the human action is denoted as Golden.
The \textit{Request} action indicates the collection of patient information, while the \textit{Inform} action indicates the provision of advice to the patient.
}
\label{fig:llm-slm}
\end{figure}

\subsection{Response Ranking Module}
% To exploit the strengths of different retrieval strategies, we incorporate the response ranking module to determine the most suitable generated response. 
Through the experimentation, we observe that LLMs tends to generate a significant number of medical terms during dialogue, resulting in more comprehensive responses. 
% But the dialogue actions performed by LLMs are often considered incorrect, leading to an overall decline in response quality. 
However, LLMs often exhibit overconfidence, as they question patients less frequently, resulting in a decrease in the overall quality of responses.
On the contrary, small language models (SLMs) that are fine-tuned on a dialogue corpus behave cautiously and tend to include a limited number of medical terms in their output. 
This phenomenon is illustrated in Figure~(\ref{fig:llm-slm}).
% Conversely, small language models (SLMs) that have been fine-tuned on the dialogue corpus can employ more accurate dialogue actions but tend to include a limited number of medical terms in its outputs.
Acknowledging the complementary nature of LLMs and SLMs, we propose using a SLM to evaluate the responses of the LLM.

Specifically, for a given sample with a dialogue history $h$, we use perplexity as the score of the response $r$ generated by the LLM. 
We compute the score using the following equation:
\begin{equation}
s(r) = - \frac{1}{l} \sum_{i=1}^l \log p(r_i \mid r_{<i}, h;\theta)
\end{equation}
Here, $l$ represents the length of response $r$, and $\theta$ denotes the model parameter of the SLM. 
%As shown in Figure~(\ref{fig:structure}), Response Ranking Module. 
This evaluation model uses an encoder-decoder architecture, where $h$ is input to the encoder side of the model, and $r_{<i}$ is input to the decoder side for calculating the generation probabilities.
We select the response with the lowest score as the system output.
\section{Automatic Evaluation Metrics} \label{section:metrics}

We observe that previous studies~\cite{varshney2023a, varshney2023b, meddialog} employ the metrics for open domain dialogue tasks that often cannot effectively measure system performance in task-oriented settings~\cite{Ji2022, sas}. 
From the results given in Figure~(\ref{fig:llm-slm}), we observe that the LLM exhibits overconfidence. 
To judge whether our approach alleviates the problem, it becomes crucial to evaluate the dialogue actions taken by the LLM.
%Additionally, evaluating the content of the conversations generated by the LLM is equally important.
It is equally important to evaluate the dialogue content generated by LLM.
Hence, we introduce two metrics that consider both intent and the usage of high-frequency medical terms.

\subsection{Intent Evaluation} 
The intent accuracy~(Int) is used to assess the conformity of the system's response to the ground-truth in terms of dialogue actions.
We train a medical dialogue intent classifier for calculating Int, and Int is calculated by the following formula:
\begin{equation}
    Int = \frac{1}{N} \sum_{i = 1}^N f(Pred_i, Golden_i)
\end{equation}
Here, $N$ represents the total number of samples. $Pred_i$ and $Golden_i$ denote the model's predicted response and the corresponding actual response, respectively. 
The function $f(\cdot, \cdot)$ evaluates the intentions, which are extracted by the aforementioned intent classifier, assigning a value of 1 if the intentions of the two responses are the same, and a value of 0 otherwise.
Appendix~\ref{sec:ic} presents the implementation details of the intent classifier.

\subsection{Medical Term Evaluation} 
We evaluate the completeness and correctness of the responses using the micro-f1 score, which measures the overlap of high-frequency medical terms between the ground-truth and the prediction.
% However, utilizing exact matches directly is deemed unreasonable in light of the non-uniqueness of treatments for diseases within the medical domain. 
We need to avoid relying solely on exact matching when measuring the term overlap. 
%due to the need to consider certain special diagnostic scenarios.
For instance, 
%in the case of some specific ailments, various physicians may prescribe different drugs, which are equivalent efficacy, to their patients. 
different doctors may prescribe different medications for some diseases that may have same effect on the patient.
% Using exact matching may lead to underestimation of the performance of models capable of generating different treatment options.
Employing exact matching alone may result in an underestimation of the performance exhibited by models that possess the ability to generate diverse treatment options.

Hence, we introduce a novel approach called Top-n Match (TnM) to address the problem.
%The definition of TnM is as follows.
% In addition to the intent assessment, we assess the conversational content in more detail by checking whether the model responses are consistent with the ground-truth to high-frequency medical terms.
% We utilized the Chinese medical high-frequency vocabulary acquired from THUOCL~\footnote{https://github.com/thunlp/THUOCL} as the term list, comprising 18,749 frequently employed terms in doctor-patient communication. 
% Initially, similar to the scheme of \citet{Li2021}, we employ word matching to identify high-frequency medical terms in system responses and golden responses, leveraging the aforementioned vocabulary. 
% Subsequently, we calculate the micro-F1 score for term matching. 
% In this scoring process, we introduce a novel approach called Top-n Match (TnM) that takes into account the similarity between terms instead of relying only on exact matches.
% The definition of TnM is as follows.
Concretely, let $T=\{t_1, t_2, ..., t_{|T|}\}$ be a set consisting of $|T|$ terms, and let $s$ be a similarity score function that satisfies $0 \le s(t_i, t_j) \le 1$ for all $t_i, t_j \in T$.
For a given term $t_i \in T$, we define $S^n_i(s, t_i) \subseteq T$ as the set of $n$ terms that are closest to $t_i$ based on the similarity function $s$.
We say that $t_i$ and $t_j$ are Top-N Match if and only if $S^n_i(s, t_i) \cap S^n_j(s, t_j) \neq \emptyset$.
TnM with different $n$ can represent the term-matching scores for different similarity levels. 
We present the f1-score results using T3M configurations.
% For our implementation, the cosine similarity of the vector is denoted by $s$, and we generate the term vectors through the utilization of the skip-gram algorithm.
Appendix~\ref{sec:te} and ~\ref{sec:tm} will give more details about the term extraction and matching.

\section{Experiments}
This section delineates the evaluation setup and the results of the proposed approach in the context of medical dialogue generation.
%, encompassing fluency, terminology correctness, and intent correctness. 
% We also demonstrate the performance through manual evaluation.
\begin{table}[t!]
\centering
%\resizebox{.95\columnwidth}{!}{
\begin{tabular}{l|l|r}
\toprule
\textbf{Dataset} & \textbf{Train / Valid / Test} & \textbf{Turn}\\
\midrule
    Meddg & 14,864 / 2,000 / 1,000 & 9.92 \\
    MedDialog & 32,723 / 3,000 / 3,000 & 4.76  \\
    KaMed & 57,754 / 3,000 / 3,000 & 11.62\\
\bottomrule
\end{tabular}
\caption{The statistics of datasets, where \textit{Turn} indicates the average number of rounds contained in each dialogue session contains.}
\label{table:dataset}
\end{table}

\begin{table*}[th]
\scalebox{0.98}{
\begin{tabular}{@{}c|cccc|cccc|cccc@{}}
\toprule
\multirow{2}{*}{\textbf{Model}} & \multicolumn{4}{c}{\textbf{Meddg}} & \multicolumn{4}{c}{\textbf{MedDialogue}} & \multicolumn{4}{c}{\textbf{KaMed}} \\ \cmidrule(l){2-13} 
                       & \textbf{R-L}  & \textbf{B-S} & \textbf{T3M}  & \textbf{INT}  & \textbf{R-L}   & \textbf{B-S}    & \textbf{T3M}   & \textbf{INT}   & \textbf{R-L} & \textbf{B-S}  & \textbf{T3M}  & \textbf{INT}  \\ \midrule
\multicolumn{13}{c}{\textit{Fine-tuning based}}                                                                                   \\ \midrule
Bart                   & 24.1   & 64.8 & 9.9  & 42.0 & 11.8    & 59.0   & 16.2  & 42.0  & 15.9  & 60.1 & 15.5 & 45.0 \\
Mars                   & 20.9   & 63.1 & 10.3 & 39.9 & 10.5    & 58.4   & 16.8  & 37.7  & 12.7  & 59.3 & 16.4 & 41.4 \\ \midrule
\multicolumn{13}{c}{\textit{LLM based}}                                                                                   \\ \midrule
Bloom                  & 14.2   & 60.3 & 10.9 & 26.3 & 10.7    & 57.2   & 16.2  & 33.0  & 11.3  & 58.7 & 16.9 & 34.1 \\
Bloomz                 & 14.5   & 58.9 & 8.5  & 24.8 & 10.0    & 57.9    & 12.4  & 31.9  & 9.7  & 57.8 & 15.0 & 32.0 \\ 
ICL Rand               & 16.4   & 61.2 & 10.6 & 29.6 & 9.9    & 57.3    & 14.8  & 36.6  & 9.3  & 56.9  & 15.0 & 36.1 \\
ICL Sbert              & 18.7   & 62.7 & 11.3 & 36.3 & 11.9    & 60.0   & 17.7  & 37.3  & 12.0  & 59.5 & 16.8 & 37.9 \\
PlugMed(ours)                & \textbf{21.1}   & \textbf{64.1} & \textbf{12.1} & \textbf{41.3} & \textbf{12.8}    & \textbf{60.4}   & \textbf{18.4}  & \textbf{37.9}  & \textbf{14.1}  & \textbf{60.1} & \textbf{18.1} & \textbf{41.7} \\ \bottomrule
\end{tabular}
}
\caption{Automatic evaluation on the Meddg, MedDialog and KaMed datasets. R-L and B-S denote Rouge-L and Bert-Score, respectively. Boldface scores indicate best results. The performance improvement of PlugMed over the baselines is significant with $p<0.05$.}
\label{table:main_result}
\end{table*}

\subsection{Datasets and Evaluation Metrics}
We conduct experiments on three large datasets for our evaluation. 
1) The \textbf{Meddg} dataset~\cite{meddg} which is collected from \textit{Doctor Chunyu}\footnote{https://www.chunyuyisheng.com/} and consists of 17,864 dialogue sessions. 
2) The \textbf{MedDialogue-CN} dataset~\cite{meddialog} which is collected from \textit{HaoDaifu}\footnote{https://haodf.com} and comprises 38,723 dialogues without any provided annotations.
3) The \textbf{KaMed} dataset~\cite{kamed} which is also collected from \textit{Doctor Chunyu}, but at a larger scale, containing 63,754 dialogue sessions.
Statistics of three datasets are presented in Table~(\ref{table:dataset}).

To evaluate the quality of generation, we employ the Rouge-L~\cite{lin-2004-rouge} and Bert-Score~\cite{zhang2019bertscore} to measure the overall similarity between the generated text and ground-truth. 
We also utilize micro-F1 scores for entity matching in T3M settings to assess medical term correctness based on the aforementioned definitions. 
Additionally, we employ the INT metric to measure the accuracy of the intended responses.
The validity analysis of each metric can be found in the Section~\ref{sec:eve_metric}.

\subsection{Implementations}
Our approach employs BLOOM as the foundation model, and to ensure reproducibility, we avoid any form of sampling and instead utilize a greedy decoding strategy. 
We generate a set of four prompts for a given sample, including an Instruction prompt (referred to as ``Vanilla'') and three In-context prompts. 
These prompts are as follows:
1) \textbf{Vanilla}: This strategy instructs the model to act as a doctor by prefixing the samples with an instruction.
2) \textbf{Global View}: This strategy involves looking for examples through a global retriever, using full context as query.
3) \textbf{Local Primary}: In this strategy, we first consider the patient's chief complaint to retrieve examples with similar symptoms. 
From these examples, a number of samples are randomly selected to generate the prompt. 
This strategy corresponds to the initial step of the local retriever's two-step search.
4) \textbf{Local Secondary}: This strategy searches for examples through a local retriever and utilizes a full two-step search. 
All examples are from the training set of the dataset.

Limited by input length, the In-context prompt contains 4 examples, while each example is limited to the last 140~($m=20, n=120$) tokens. 
When constructing the symptom index using the K-Means algorithm, we set the number of cluster centers to 100 and iterate 20 times.
During response ranking, BART serves as the scoring model. 
We conduct the experiments using PyTorch\footnote{https://pytorch.org/} and Huggingface Inference API\footnote{https://huggingface.co/docs/api-inference/index}.

\subsection{Baselines}
\paragraph{Fine-tuning Based Baselines.} 
These baselines utilize small language models as the backbone, trained on the aforementioned datasets for the medical dialogue generation task, which include the following models:
1) \textbf{Bart}~\cite{bart}, a well-known encoder-decoder model that is recognized for its text generation capabilities.
2) \textbf{Mars}~\cite{mars}, an advanced model explicitly crafted for the MultiWoZ~\cite{multiwoz} dataset, renowned for its exceptional performance in addressing the Task-Oriented Dialogue (TOD) task. 
% We incorporate the model into our specific task and prioritized the acquisition of medical terminology through the training of dialogues.
We migrated Mars into our tasks and trained it to focus on the generation of medical terms.
Appendix~\ref{sec:mars} gives the further information on the migration process.

\paragraph{LLM Baselines.}
These baselines employ large language models to generate dialogue responses. Our comparison targets include:
1) \textbf{Bloom}~\cite{bloom}, a widely used open-source multilingual language model with 176 billion parameters.
2) \textbf{Bloomz}~\cite{bloom}, an instruction-tuned model derived from Bloom and specializes in zero-shot tasks.
In addition to these models that use Instruct-prompt as input, we also include two baselines that utilize In-context prompts as input:
1) \textbf{ICL Rand}, which selects a set of dialogue examples randomly to construct the prompt.
2) \textbf{ICL Sbert}, which utilizes Sbert~\cite{sentence-bert} to encode the dialogue history and find the closest examples to the given sample.
Both baselines use Bloom as the fundamental model.

\begin{table*}[ht]
\centering
\begin{tabular}{@{}l|l|cccc|cccc@{}}
\toprule
\multirow{2}{*}{Dataset}     & \multirow{2}{*}{Settings} & \multicolumn{4}{c|}{Bloom} & \multicolumn{4}{c}{ChatGPT} \\ \cmidrule(l){3-10} 
                             &                           & R-L   & B-S  & T3M  & INT  & R-L   & B-S   & T3M  & INT  \\ \midrule
\multirow{5}{*}{Meddg}       & Vanilla                   & 14.2  & 60.3 & 10.9 & 26.3 & 6.7   & 56.1  & 10.5 & 23.4 \\
                             & Global View               & 18.7  & 62.7 & 11.3 & 36.3 & 12.3  & 60.0  & 12.7 & 33.8 \\
                             & Local Primary             & 18.7  & 62.5 & 10.7 & 34.6 & 11.6  & 56.1  & 12.3 & 30.7 \\
                             & Local Secondary           & 18.0  & 62.3 & 11.2 & 34.9 & 11.7  & 59.7  & 12.1 & 30.3 \\
                             & +Ranking (Ours)           & 21.1  & 64.1 & 12.1 & 41.3 & 12.9  & 60.4  & 12.9 & 34.4 \\ \midrule
\multirow{5}{*}{MedDialogue} & Vanilla                   & 10.7  & 57.2 & 16.2 & 33.0 & 7.5   & 56.5  & 16.5 & 40.5 \\
                             & Global View               & 11.9  & 60.0 & 17.7 & 37.3 & 10.1  & 58.6  & 17.8 & 38.8 \\
                             & Local Primary             & 11.0  & 59.0 & 15.5 & 36.6 & 10.0  & 58.5  & 18.3 & 38.2 \\
                             & Local Secondary           & 12.0  & 60.0 & 16.7 & 36.1 & 9.9   & 58.4  & 18.0 & 39.1 \\
                             & +Ranking (Ours)           & 12.8  & 60.4 & 18.4 & 37.9 & 10.1  & 58.6  & 18.5 & 40.2 \\ \midrule
\multirow{5}{*}{KaMed}       & Vanilla                   & 11.3  & 58.7 & 16.9 & 34.1 & 6.1   & 54.9  & 14.9 & 36.2 \\
                             & Global View               & 12.0  & 59.5 & 16.8 & 37.9 & 8.9   & 57.5  & 17.1 & 39.8 \\
                             & Local Primary             & 12.0  & 59.5 & 17.1 & 39.4 & 8.6   & 57.2  & 17.0 & 38.9 \\
                             & Local Secondary           & 12.1  & 59.4 & 17.7 & 36.7 & 8.4   & 57.2  & 17.1 & 38.7 \\
                             & +Ranking (Ours)           & 14.1  & 60.1 & 18.1 & 41.7 & 9.1   & 57.6  & 17.1 & 40.5 \\ \bottomrule
\end{tabular}
\caption{Ablation studies on the Meddg, MedDialog and KaMed datasets. \textit{Vanilla}, \textit{Global view}, \textit{Local primary} and \textit{Local secondary} represent the four prompt generation strategies respectively, and \textit{+Ranking} represents the results of the selection based on the four strategies using the RR module.}
\label{table:ab_study}
\end{table*}

\subsection{Overall Performance}
Table~(\ref{table:main_result}) present the automatic evaluation results on Meddg, MedDialog, and KaMed. 
Remarkably, PlugMed consistently attains the top-ranking positions across a majority of the metrics and achieves best performance in T3M, outperforming even the strongest baseline. 
Meanwhile, among all baselines leveraging the LLM, PlugMed generates responses with the most reasonable intent.

Our analysis uncovers interesting observations. 
Firstly, we discovered that fine-tuned small models tend to have higher INT scores but lower term matching-related scores. 
This suggests that while these models excel in emulating the dialogue actions of doctors, they often struggle to generate appropriate medical terminology due to their limited medical knowledge.
Secondly, we observed that Bloomz performed worse than Bloom, indicating that the instruct-tuning process may compromise the diagnostic capability of the model. 
More details can be found in Section~\ref{sec:gc}.
% Additionally, we found that the output of BLOOMZ is less affected by the examples compared to BLOOM (see Appendix~\ref{sec:bloomz} for more details), suggesting that the instruct-tuning may also compromise the ICL capabilities of the model.

\subsection{Ablation Study}
To investigate the influence of different prompt generation strategies on system responses and the efficacy of the RR module, we conduct ablation experiments. 
The experimental results, shown in Table~(\ref{table:ab_study}), indicate that both the global and local view effectively enhance the quality of the LLM output. 
% Notably, the global view strategy outperforms the local view strategy, probably due to its capability to consider more diverse information beyond solely focusing on patient symptoms. 
Moreover, we observe a substantial enhancement in the quality of system responses due to the integration of the RR module. 
This observation underscores the efficacy of the fine-tuned SLM in effectively evaluating the LLM's output. 

Furthermore, we observe that Global View demonstrates comparable performance to PlugMed when assessed using the MedDialogue dataset. 
However, PlugMed significantly surpasses Global View on the Meddg and KaMed datasets. 
We attribute this discrepancy to the MedDialogue dataset's characteristic of having shorter dialog histories, averaging only 4.76 rounds. 
Conversely, Global View's retrieval effectiveness diminishes as the samples' length increases, as elaborated in Section~\ref{sec:bi}, making it less effective than PlugMed on the other two datasets. 
This underscores the synergistic relationship between Global View and Local View, highlighting their complementary strengths.

\begin{figure}[th]
\centering
\includegraphics[width=0.48\textwidth]{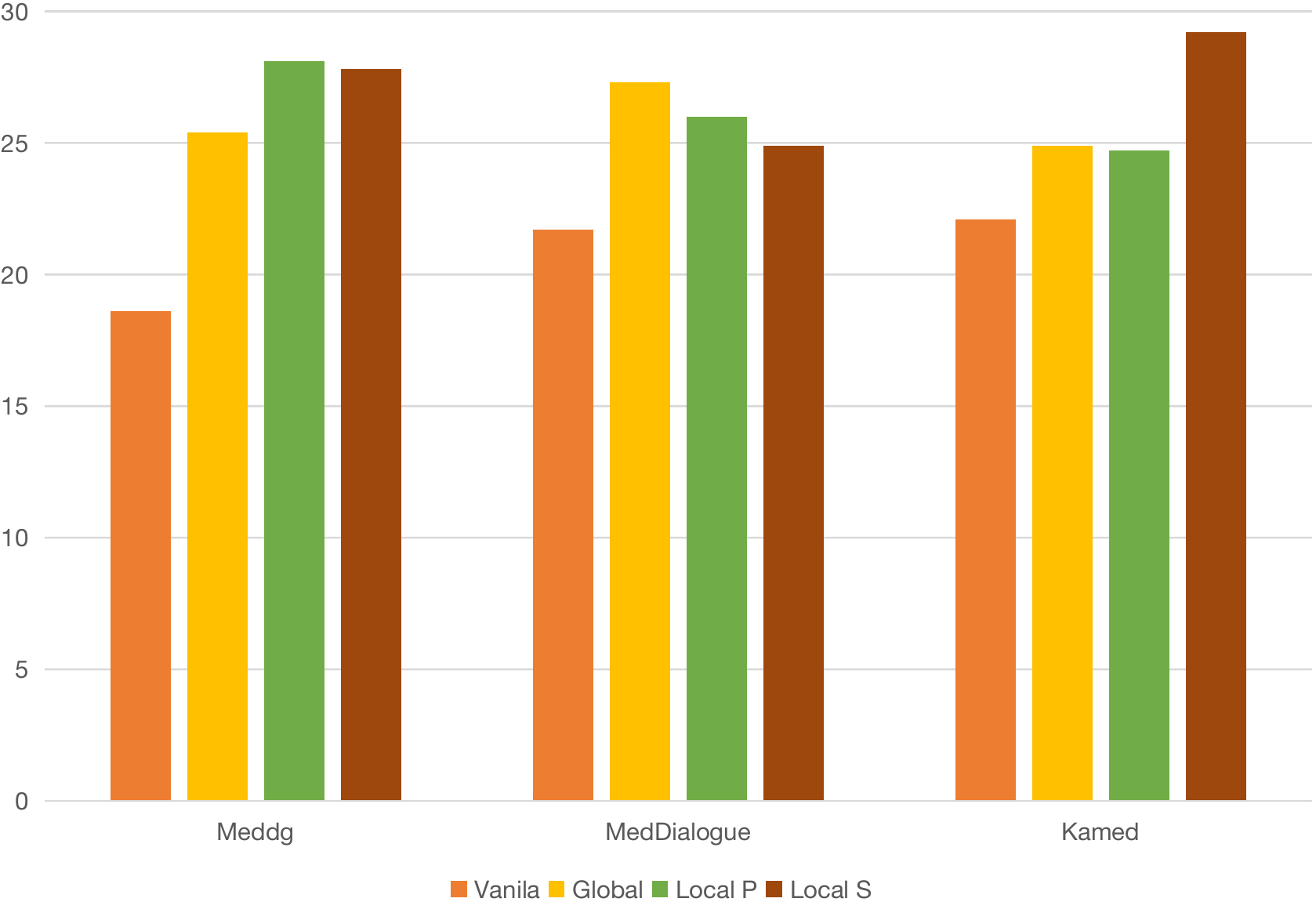} 
\caption{The evaluation results of the RR module for four strategies. The vertical axis represents the proportion of the strategy get the best score. \textit{Local P} and \textit{Local S} correspond to Local Primary and Local Secondary, respectively.}
\label{fig:compare}
\end{figure}

To delve deeper into the contributions of these strategies to the outcomes, we utilize the RR module to rank the responses corresponding to each strategy, documenting the percentage of times each strategy achieves the top rank. 
The results are illustrated in Figure~(\ref{fig:compare}). 
Based on the findings depicted in the figure, we can infer that the responses generated by the Vanilla strategy exhibit significantly inferior quality compared to those produced by the other three strategies. 
Furthermore, the probability of being selected by the remaining three strategies is approximately equal, indicating their complementary nature.

\subsection{Analysis of Generalization Capacity} \label{sec:gc}
We conducted experiments using ChatGPT and Bloomz to assess the effectiveness of our approach in terms of generalization. 
The results, as presented in Table~(\ref{table:ab_study}) for ChatGPT, revealed that our approach successfully enhances ChatGPT's performance in completing medical conversations. 
However, it was observed that ChatGPT's capability for multi-round conversations lags behind that of Bloom. 
This discrepancy can be attributed to the safety protocols integrated into ChatGPT by OpenAI. 
Specifically, ChatGPT often advises patients to seek professional assistance instead of providing diagnoses, which affects its multi-round conversation potential. 
Conversely, we noticed that our approach has minimal impact on Bloomz, as detailed in Appendix~\ref{sec:bloomz}. 
This disparity can be attributed to Bloomz's training dataset, bigscience/xP3~\footnote{https://huggingface.co/datasets/bigscience/xP3}, which primarily comprises single-round Q\&A tasks, making it less adaptable to multi-round conversations. 
To summarize, the generalization ability of our method is influenced by the model's pre-training task, and improving our method's effectiveness entails pre-training models based on multi-round dialogues and reducing safety interventions.

\subsection{Human Evaluation} \label{sec:hum}
\begin{table*}[th]
\small
\begin{tabularx}{\textwidth}{l|X}
\toprule
\textbf{Check Item}       & \textbf{Content}                                                                                                          \\ \midrule
Role Consistency & The response should exhibits a doctor-like response style and utilizes natural language.                                       \\
Empathy          & The response should demonstrate an understanding of the user's intents and employ a friendly tone.                      \\
Correctness      & The responses must adhere to common medical sense and be evidence-based.                                  \\
Necessity        & The response should provide assistance in advancing the diagnosis or satisfying the patient's curiosity.                 \\
Richness         & The responses should incorporate additional relevant medical information to significantly enhance the patient experience. \\
\bottomrule
\end{tabularx}
\caption{Check items for human evaluation.}
\label{table:manual_eval}
\end{table*}

\begin{table*}[ht]
\centering
\begin{tabular}{@{}l|cccccccc@{}}
\toprule
Model       & Human & BLEU  & R-L  & B-S  & T1M  & T3M  & T5M  & INT  \\ \midrule
Golden      & 90.4  & -     & -    & -    & -    & -    & -    & -    \\ \midrule
Bart        & 58.9  & 4.1   & 22.1 & 63.0 & 8.3  & 11.6 & 12.3 & 40.9 \\
Mars        & 44.6  & 5.4   & 19.2 & 62.3 & 9.1  & 12.8 & 13.9 & 33.2 \\
Bloom       & 46.7  & 3.8   & 13.8 & 59.6 & 7.4  & 8.8  & 10.9 & 24.0 \\
Bloomz      & 49.8  & 3.3   & 13.7 & 59.5 & 3.6  & 5.8  & 7.3  & 28.3 \\
Bloom-Rand  & 52.9  & 3.9   & 15.4 & 61.0 & 7.3  & 13.2 & 14.8 & 30.1 \\
Bloom-Sbert & 57.5  & 4.3   & 16.5 & 61.9 & 9.9  & 14.0 & 16.2 & 33.1 \\
PlugMed     & 61.0    & 4.1   & 19.3 & 62.8 & 8.2  & 13.0 & 14.6 & 37.3 \\ \midrule
Pearson(\%) & -     & -22.7 & 48.9 & 59.7 & 27.1 & 42.8 & 41.2 & 71.3 \\ \bottomrule
\end{tabular}
\caption{
We evaluated three datasets by human ratings.
For evaluation purposes, we randomly selected 100 samples from each dataset. 
The average score for each sample was calculated based on the assessments of three physicians.
}
\label{table:hum}
\end{table*}

We manually evaluate seven selected dialogue models to conduct a comprehensive comparison. 
We ensure a thorough analysis by randomly selecting 100 samples from each dataset. 
Each sample is examined by three physicians who assign scores based on the criteria outlined in Table~(\ref{table:manual_eval}). 
The evaluators check the responses in the following order: Role Consistency, Empathy, Correctness, Necessity, and Richness. 
This order also corresponds to the importance of the check items, and subsequent items only had test significance if the model satisfied the previous check item. 
Therefore, we apply the following rules for scoring. 
If a target response successfully pass a particular test, it receives a score of 20. 
However, if the target response does not pass a specific test, it is assigned a score of 0 for all subsequent check items.

We compute the average score of each model on the dataset and summarize the evaluation results in Table~(\ref{table:hum}). 
Our analysis reveals that PlugMed exhibits the highest performance based on human evaluations.
However, there still exists a significant gap between the responses generated by the models and the human responses, indicating that the models have not fully grasped the diagnostic capabilities. 
Additionally, we observe that BART performs well in the manual evaluation, primarily because we give a very low priority to richness.
Appendix~\ref{sec:case_study} provides a case study for illustrating this finding.

\subsection{Analysis of Evaluation Metrics} \label{sec:eve_metric}
Within this section, we have included an evaluation of the metric's reliability. 
To gauge the quality of these metrics, we employed the Pearson correlation coefficient to quantify their alignment with human evaluations. 
A score approaching 1 indicates a stronger metric performance. 
The corresponding outcomes can be found in Table~(\ref{table:hum}).

Our examination indicates that the INT metric surpasses all others in performance, with Bert-Score coming in a close second. 
These two metrics exhibit stronger correlations with human assessment, indicating the superiority of a semantically based evaluation. 
Simultaneously, this outcome implies that the model's intent should align closely with the corpus.
Lower scores for BLEU~\cite{bleu} and higher scores for Rouge-L suggest a preference among individuals for more comprehensive model responses over precision-oriented ones.
Lastly, based on the outcomes of T1M~(exact match), T3M, and T5M, it becomes evident that factoring in term similarity is imperative when calculating term matching scores.

\section{Related Work}
\subsection{Medical Dialogue Systems}
According to system architecture, medical dialogue systems can be of two types: the pipeline and the end-to-end systems~\cite{valizadeh-parde-2022-ai}. 
The pipeline systems usually involve four steps: the natural language understanding, the dialogue state tracking, the dialogue action generation, and the natural language generation.
%There are several attempts to enable the medical dialogue in pipeline way.
\citet{wei-etal-2018-task, xia2020generative} propose to learn the dialogue policies for automated diagnosis using reinforcement learning.
In other studies, \citet{lin-etal-2019-enhancing} proposes to model the associations between symptoms by constructing a symptom graph, aiming to enhance symptom diagnosis performance. 
\citet{Li2021} proposes to use symptoms and diseases as keys to generate responses instead of dialogue states and actions based on a knowledge graph.

The end-to-end models, usually a sequence-to-sequence architecture~\cite{seq2seq}, first attracted attention. 
The fine-tuning pre-trained models, such as GPT-2~\cite{Radford2019LanguageMA}, have been proven effective in task-based dialogue scenarios, as demonstrated by \citet{Su2021, ubar, Tods}.
BioBERT~\cite{biobert} and BioGPT~\cite{biogpt} try to improve the performance by employing pre-training on medical corpora. 
% Another solution proposed by \citet{varshney2023a, varshney2023b,Tang2022} involves using the Unified Medical Language System (UMLS) to explicitly incorporate knowledge during dialogue generation.
\citet{varshney2023a, varshney2023b,Tang2022} propose explicitly using the Unified Medical Language System (UMLS) to incorporate knowledge in the dialogue generation process.
% In summary, previous work has focused on domain knowledge enhancement and learning domain strategies.

In summary, existing studies mainly emphasize the role of knowledge enhancement based on small models. 
In contrast, our work is conducted on knowledge-rich LLMs, emphasizing the enhancement of dialogue strategies.

\subsection{ICL for Dialogue}
As LLMs continue to advance, ICL has emerged as a new paradigm in natural language processing. 
The exploration of ICL's ability to evaluate and infer LLMs has become a prominent trend~\cite{Dong2022}.
Some studies have tried to apply ICL in dialogue generation. 
\citet{Roy2023} proposes a two-stage style transfer framework to leverage ICL for dialogue style transfer. 
\citet{Meade2023, lee-etal-2022-gpt} employ a retrieval-based framework to mitigate bias and toxicity in chatbot-generated responses, guiding the model towards safer and more responsible dialogue. 
% \citet{lee-etal-2022-gpt} proposes a similar approach to encourage LLMs to generate compassionate dialogue.
%ICL has also demonstrated success in dialogue state tracking~(DST) tasks. 
\citet{Hu2022, Chen2023} propose a method for long dialogue compression, enabling each prompt to contain more examples and improving dialogue state tracking performance.
ICL has also been utilized for unsupervised generation of dialogue data in certain contexts~\cite{Dialogic}, expanding the potential applications of this approach.
% Overall, the adoption of ICL in dialogue generation has shown promising results across various aspects, including style transfer, reducing bias and toxicity, promoting compassionate dialogue, dialogue state tracking, and unsupervised generation of dialogue data.

Overall, the aforementioned works concentrate on either example retrieval or dialogue compression. 
In contrast, our work combines the two techniques for integrating multiple retrieval strategies.
\section{Conclusion}
% In this study, we use in-context learning to develop a patient-centered medical dialogue model. 
In this paper, we use in-context learning to develop a patient-centered medical dialogue model. 
% To accomplish this objective, we introduce a Prompt Generation module capable of generating LLM input from both global and local views. 
To this end, we introduce a Prompt Generation module capable of generating LLM input from both global and local views. 
Additionally, we construct a Response Ranking module using a supervised trained small model to filter the LLM output. 
Experimental results indicate that the responses generated by PlugMed exhibit a greater inclusion of comprehensive medical terms and PlugMed yields more accurate dialogue intents than other large language model based methods.

\section*{Limitations}
Based on human evaluation,  we have identified shortcomings in PlugMed's diagnostic efficiency. 
% This indicates that the model is unable to swiftly identify the patient's disease, which can result in an increase in average conversational discourse. 
This suggests that PlugMed has difficulty rapidly identifying the patient's disease, which may lead to an increase in average conversational discourse and harm the patient's experience. 
Our future work will prioritize addressing this issue.
\section{Acknowledgement}
Our work is supported by the National Key Research and Development Program of China (Project Number: 2020AAA0109400). We kindly appreciate all the researchers who provide valuable insights, discussions, and comments on this work.

\bibliography{emnlp2023}
\bibliographystyle{emnlp2023}

\appendix

\section{Details of Automatic Evaluation}
\label{sec:detail}

\subsection{Intent Classifier} \label{sec:ic}
To compute Int, we employ the IMCS-IR dataset~\cite{icms} and utilize Roberta-large~\cite{roberta} as the backbone model for training the classifier. 
To process a response, we concatenate it with the dialogue history and input the combined sequence into the model for classification. 
The input format is structured as follows:
\begin{quote}
<s> [dialogue history] </s> [response]
\end{quote}
We extract the hidden layer vector of the `<s>` token as the embedding representation for the response, which is subsequently classified using a two-layer neural network with a hidden dimension of 768. 
The intentions considered are presented in Table~(\ref{table:intent}), and the model achieves an accuracy of \textbf{0.86} on the validation set.

\begin{table}[h]
\scalebox{0.85}{
\begin{tabular}{l|l}
\toprule
\textbf{Action Type} & \textbf{Target}                             \\
\midrule
Request     & Symptom                            \\
Request     & Etiology                           \\
Request     & Basic Information                  \\
Request     & Existing Examination and Treatment \\
Inform      & Drug Recommendation                \\
Inform      & Medical Advice                     \\
Inform      & Precautions                        \\
Inform      & Diagnose                           \\
Other       & Other                              \\
\bottomrule
\end{tabular}
}
\caption{The intents of the doctor.}
\label{table:intent}
\end{table}

\begin{table*}[ht]
\small
\begin{tabularx}{\textwidth}{l|X|l|X}
\toprule
Role    & Dialogue History                                                                                                                                                                                                                  & Meddg    & Ours                              \\
\midrule
Patient & How to treat colitis, the doctor prescribed rehabilitation liquid and mesalachin granules, please read the bc sheet (Female, 35)                                                                                                  &             &                                   \\
Doctor  & What are your current symptoms? Have you had a gastroscopy?                                                                                                                                                                       & Gastroscopy & Gastroscopy                       \\
Patient & I feel like I have diarrhea, but I can't get it out, and my stomach doesn't feel good, I don't know if it's serious, I don't have a gastroscopy.                                                                                  &             &                                   \\
Doctor  & The medication prescribed by the doctor is symptomatic, just keep taking it. There may be ulcers in the intestines. Usually, you should also pay attention to your diet, avoid spicy stimulation, quit smoking and limit alcohol. &             & Medicine, Intestinal, Ulcer, Diet \\
Patient & Is it necessary to do a gastroscopy?                                                                                                                                                                                              &             &                                   \\
Doctor  & Because there is an ulcer in the intestine, the stomach should also be checked. But it is okay to take medicine first.                                                                                                            &             & Intestine, Ulcer, Stomach, Check  \\
Patient & How long does this mesalachin need to take?                                                                                                                                                                                       &             &                                   \\
Doctor  & The duration of medication is determined by the symptoms. When you feel well, you can stop the medication. Pay attention to your diet during the day.                                                                             &             & Symptoms, Duration of medication, Diet                \\
\bottomrule
\end{tabularx}
\caption{A comparison of the entities provided by Meddg with the terms we extracted. It can be observed that our extracted terms can better cover the content of the conversation.}
\label{table:compare}
\end{table*}

\begin{figure*}[th]
\centering
\includegraphics[width=0.98\textwidth]{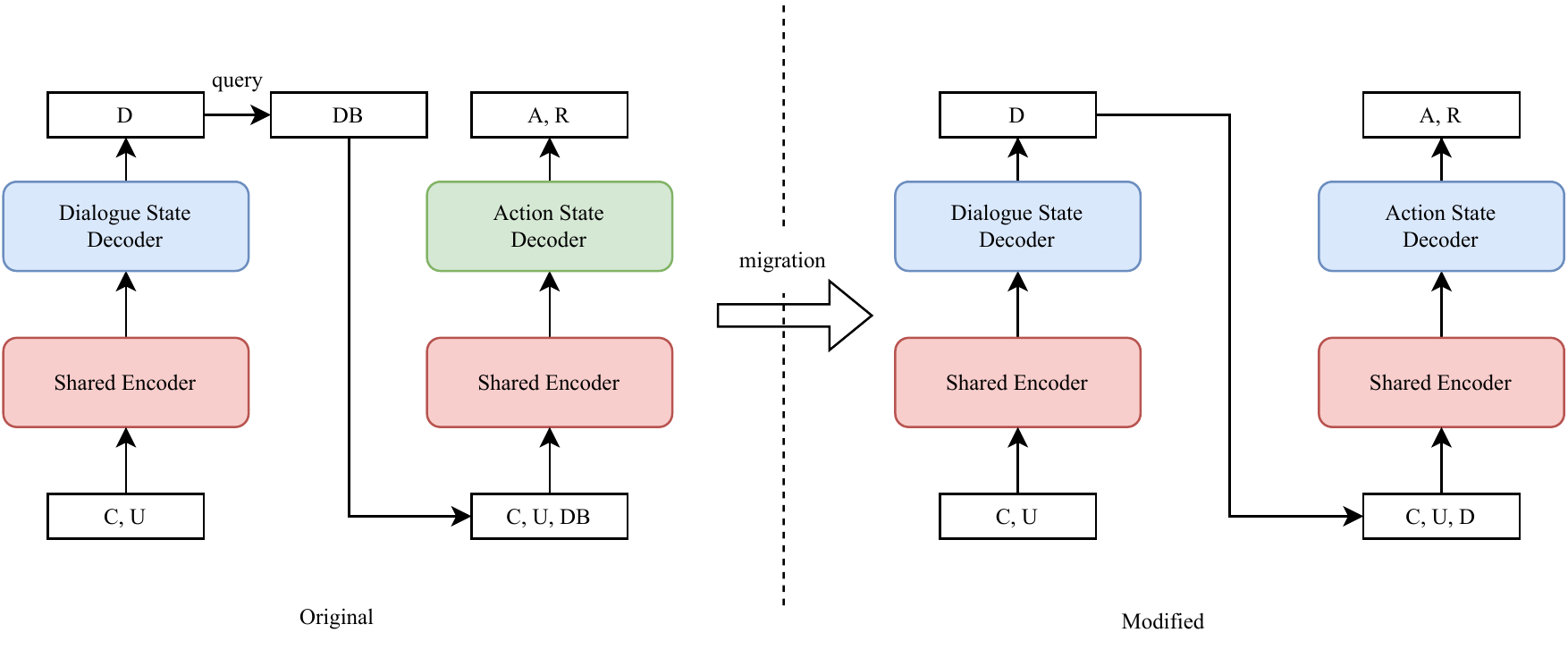} 
\caption{Migration methods for Mars models.}
\label{fig:mars}
\end{figure*}

\subsection{Term Extraction} \label{sec:te}
Following the implementation outlined in \cite{Li2021}, we employ word matching to extract high-frequency terms from responses. 
To accomplish this, we initially construct a glossary of medical high-frequency words. 
We then appropriately expand the glossary to ensure a satisfactory hit rate for term matching. 
Our approach is detailed as follows.

\paragraph{Initialization of the glossary.}
We utilized the Chinese medical high-frequency vocabulary acquired from THUOCL\footnote{https://github.com/thunlp/THUOCL} as the initial glossary, comprising 18,749 frequently employed terms in doctor-patient communication. 

\paragraph{Glossary expansion.}
We utilize the skip-gram algorithm to expand the list of high-frequency terms. 
Our approach involves using the skip-gram algorithm to discover synonyms of the initial vocabulary and incorporate them into the vocabulary. 
This method helps us identify aliases and common names of the terms and facilitates the calculation of the TnM f1-score by providing term similarity information.
To accomplish this, we train skip-gram word embeddings using the CMCQA~\cite{CMCQA} corpus, which consists of 1.3 million complete conversations, 19.83 million sentences, and 650 million tokens. 
Our skip-gram algorithm employs a word vector dimension of 300, a window size of 5, a sub-sample ratio of 3, and undergoes 8 rounds of training.
Using the obtained word vectors, we enrich the high-frequency terminology list by adding the 10 synonyms with the closest semantic meanings for each term. 
Subsequently, we save the term vectors obtained from the training to facilitate the calculation of similarity scores.

\paragraph{Discussion.}
It is worth to note that Meddg~\cite{meddg} and a few other datasets propose similar evaluation metrics based on medical entity matching. 
This raises the question of why we are considering medical terminology matching instead of medical entity matching. 
We chose terminology matching instead of entity matching because we observe several issues. 

Firstly, most medical named entity recognition (NER) datasets~\cite{Zan2021,2020CMeIE} focus on recognizing a limited range of entity types, which often excludes entities crucial to the consultation process, such as etiology related to eating habits. 
Secondly, since our target audience is patients without medical expertise, the dialogue content incorporates numerous colloquial words. 
This poses challenges for NER models that are primarily designed for professional terminology. 
Consequently, the traditional NER model can only identify a limited number of entities within the dataset. 
For example, in the report of Meddg, an average sentence contains only 0.56 entities, which hampers the evaluation of the dialogue content. 

Therefore, we believe that evaluating high-frequency medical terms would be a more logical approach. 
The glossary we selected offering an 81.9\% coverage of the labeled entity classes in Meddg and featuring an average of 2.13 entities per utterance. 
Table~(\ref{table:compare}) illustrates the comparison between Meddg's original entity extraction and our improved term extraction, clearly showing that our extracted terms better cover the dialogue content.

\subsection{Term Matching} \label{sec:tm}
Let $A_i = \{a^i_1, .., a^i_n\}$ represent the set of $n$ terms in the system response $i$, and $B_i = \{b^i_1, ..., b^i_m\}$ represent the set of $m$ terms in the standard response $i$.
We use the cosine similarity of the skip-gram vector corresponding to the term as the required similarity score for Top-n Match.
To calculate the f1-score, we define the following term types:
\begin{itemize}
    \item Truth-Positive (TP): For any $a^i_u \in A_i$, if there exists $b^i_v \in B_i$ Top-n Matches with it, then we classify $a^i_u$ as a TP term.
    \item False-Positive (FP): If $a^i_u$ is not a TP term, it is classified as an FP term.
    \item False-Negative (FN): If $b^i_v$ does not top-n match with any $a^i_u \in A_i$, we classify $b^i_v$ as an FN term.
\end{itemize}
Then the calculation formula for micro-f1 is as follows, where N denotes the number of samples, and i denotes the index of the samples.
\begin{gather}
P = \frac{\sum_i^N TP_i}{\sum_i TP_i + \sum_i FP_i} \\
R = \frac{\sum_i^N TP_i}{\sum_i TP_i + \sum_i FN_i} \\
F = \frac{2PR}{P + R}
\end{gather}

\section{Implementation of Mars}
\label{sec:mars}
The overview of the migration process are shown in Figure~(\ref{fig:mars}).
Mars~\cite{mars} incorporates two decoders, namely the Dialogue State Decoder and the Action State Decoder, which both utilize a Shared Encoder.
The original workflow of Mars is as follows:
1) Utilize the Shared Encoder to encode the context $C$ and user input $U$, and employ the Dialogue State Decoder to decode the dialogue state $D$.
2) Employ the dialogue state to retrieve the corresponding entity $DB$ from the database.
3) Utilize Shared Encoder to encode $C$, $U$, and $DB$, and employ the Action State Decoder to decode the dialogue action $A$ and dialogue response $R$.

Since our task does not involve interaction with the database, we omit the second step but employ $D$ instead of $DB$.
Considering that the dataset lacks labeled dialogue states and actions, we adopt the approach of \citet{Li2021} and utilize high-frequency medical vocabulary found in sentences as the states and actions.
Our subsequent experiments show that this approach facilitates the model to generate more medical terms while improving the fluency of the output.
We keep the other settings the same as the original configuration.

\section{ICL on Bloomz}
\label{sec:bloomz}
This section explores the adaptability of Bloomz to In-Context Learning. 
We observed that the impact of ICL on BLoomz was minimal. 
Adjusting the order of the examples or changing the number of examples did not result in significant changes in the model's performance. 
To illustrate this phenomenon, we provide the details of the following experiment.

\subsection{Experimental Setup}
In order to evaluate the impact of In-Context Learning (ICL) on the performance of Bloom and Bloomz, we conducted experiments on the validation set of the KaMed dataset. 
For each sample in the validation set, we generated 10 prompts, where each prompt consisted of a random number of randomly selected examples. The maximum number of examples included in a prompt was limited to less than 4. 
Using these 10 prompts, we made predictions using Bloom and Bloomz models respectively. 
We then count the number of non-repeats of the system's responses.

\begin{figure}[ht]
\centering
\includegraphics[width=0.48\textwidth]{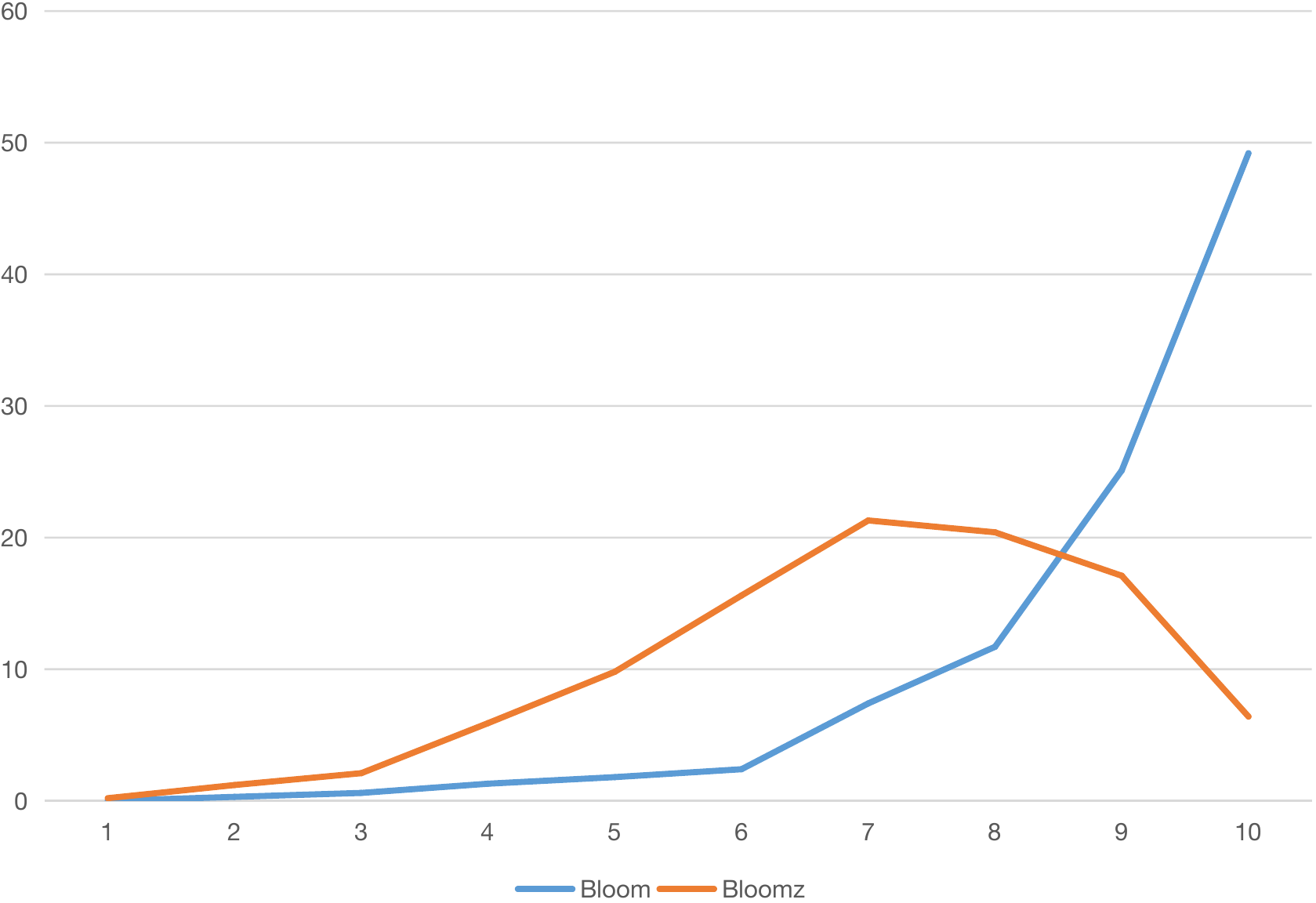} 
\caption{Diversity comparison of Bloom output and Bloomz output.}
\label{fig:comp}
\end{figure}

\subsection{Experimental Result}
The test results are presented in Figure~(\ref{fig:comp}). 
The horizontal axis of the figure represents the number of unique outputs obtained from the 10 prompts per sample, while the vertical axis represents the percentage of samples in the dataset that have that specific number of unique outputs.
From the figure, we can observe that Bloom generates almost different answers for ICL prompts containing different examples. 
In contrast, Bloomz generates more duplicate answers, indicating that Bloomz is less influenced by the sample.
The observation that Bloomz's dialogue strategy is less malleable, suggests that the Instruct-tuning process may compromise the model's In-Context Learning capabilities on the medical dialogue generation task.

\begin{table*}[th]
\small
\begin{tabularx}{\textwidth}{l|X}
\toprule
\multirow{3}{*}{History} & U: The baby has a cold for the sixth day, bronchitis, phlegm, cough, and runny nose.                                                                                                                                                                                                                                      \\
                         & U: I went to the hospital of traditional Chinese medicine and prescribed Chinese medicine without decoction.                                                                                                                                                                                                              \\
                         & U: After taking it for three days, it didn’t seem to work. What should I do (male, 13 months)                                                                                                                                                                                                                             \\ \midrule
Golden                   & Hello! Does your baby have other symptoms? Such as fever?                                                                                                                                                                                                                                                                 \\ \midrule
BART                     & Hello, is your baby coughing badly?                                                                                                                                                                                                                                                                                       \\
Mars                     & The baby is bronchial sinusitis, bronchitis, cough, runny nose.                                                                                                                                                                                                                                                           \\
BLOOM                    & Your baby needs timely medical treatment. The child's current symptoms are obviously suppurative tonsillitis. In addition to taking anti-inflammatory treatment for this disease, it is recommended to perform tonsil removal surgery at the same time.                                                                                                                                                                                                                                            \\
BLOOMZ                   & The baby is still young, so it is recommended to take western medicine. Chinese medicine has too many side effects on the baby, and the effect is not good.                                                                                                                                                               \\
ICL RAND               & If your child coughs and has phlegm, you can take some phlegm-reducing medicines, such as Mucosolvan and keep warm to avoid catching cold.                                                                                                                                           \\
ICL SBERT                & The baby is now coughing and has phlegm. It is considered to be bronchitis. You can continue to take traditional Chinese medicine, or you can take cephalosporins, combined with atomization treatment.                                                                                                                   \\
PlugMed                  & Hello, does your baby have a fever now?         \\
\bottomrule
\end{tabularx}
\caption{One case extracted from KaMed. Note that these dialogues are originally in Chinese, the English version may not get the same responses.}
\label{table:case_study}
\end{table*}

\section{Case Study}
\label{sec:case_study}
We selected a representative sample from the KaMed dataset to facilitate an interpretable comparison of different models. The corresponding results are presented in Table~(\ref{table:case_study}).
In the scenario depicted within this sample, the patient's mother provided limited information to the doctor, which is insufficient to support an accurate diagnosis. 
Consequently, the doctor made the decision to gather more necessary symptom information.

Our observations reveal that the Bart model exhibit an awareness of the information gap.
However, Bart inquire about a known symptom, indicating that the small model lacks certain medical common sense and doesn't know what information to collect.
Mars places greater emphasis on generating pertinent medical terms. 
In the absence of other information, Mars choose to reiterate the patient's words in an attempt to enhance the likelihood of a terminological hit.
Unfortunately, this approach result in the generation of responses of lower quality.
The LLM produce lengthier responses, while ICL Rand and ICL Sbert attempt to provide a diagnosis directly. 
On the other hand, Bloom and Bloomz generate false and biased utterances, respectively, indicating a dearth of diagnostic strategies within these larger models.
In contrast, PlugMed generate responses similar to the ground-turth, thus suggesting the effectiveness of our proposed method.

\section{Prompt Format} \label{sec:prompt}
\begin{figure*}[t]
\centering
\includegraphics[width=0.98\textwidth]{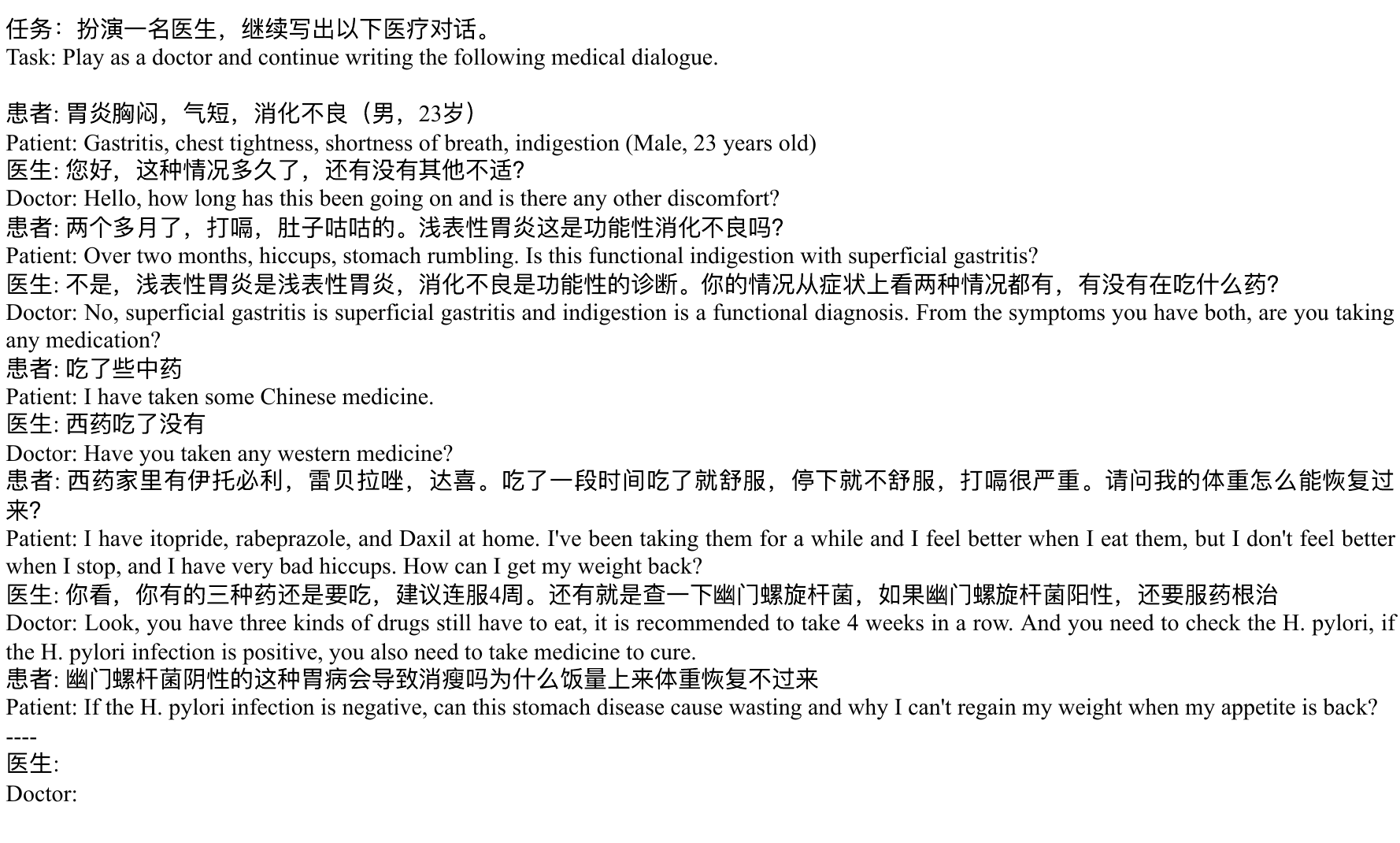} 
\caption{An example of the Instruct prompt.}
\label{fig:ins_p}
\end{figure*}

\begin{figure*}[t]
\centering
\includegraphics[width=0.98\textwidth]{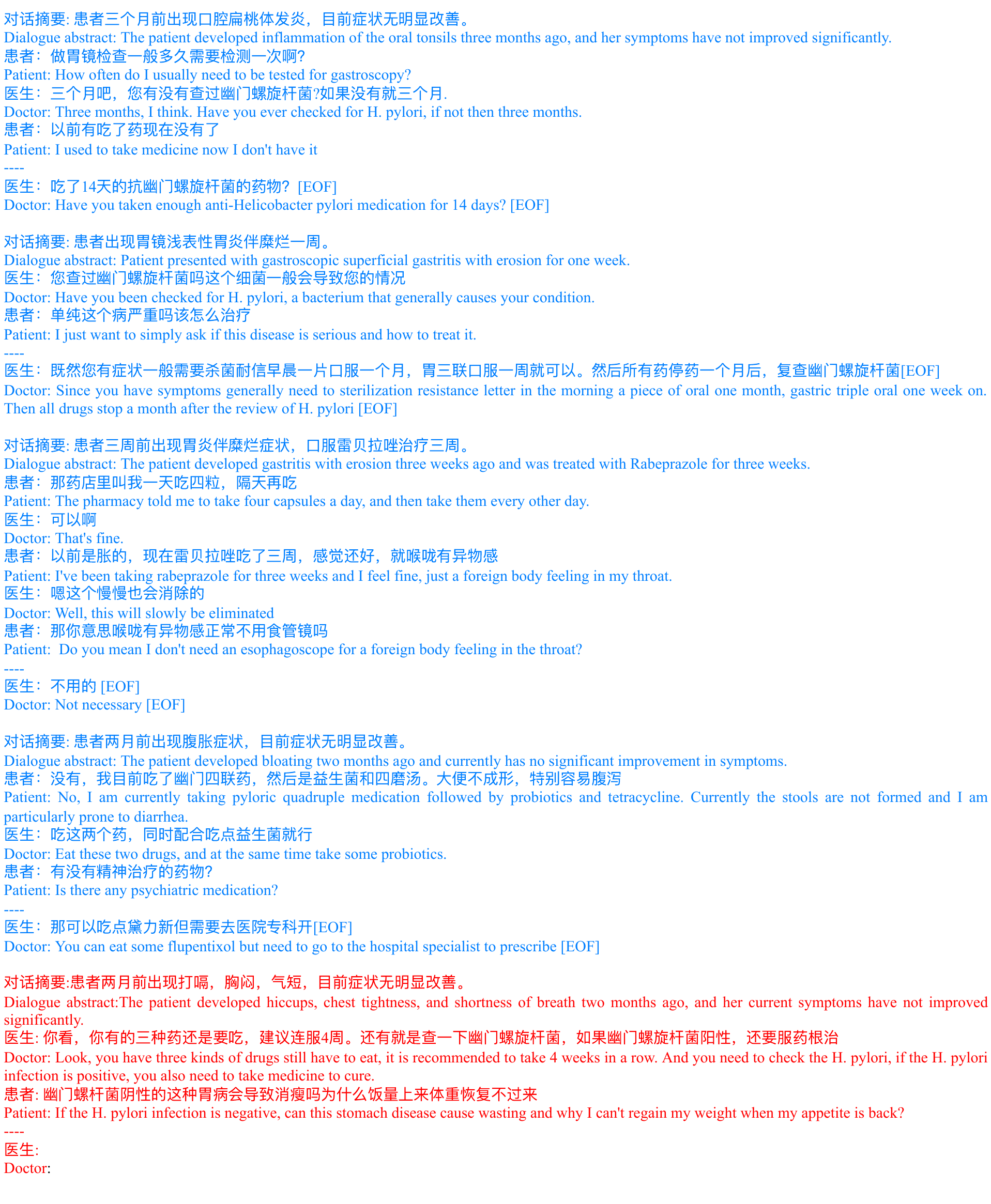} 
\caption{An example of In-context prompt. Blue text indicates examples and red text indicates the sample.}
\label{fig:icl_p}
\end{figure*}

Our experiments employ two prompts: the Instruct prompt, shown in Figure~(\ref{fig:ins_p}), and the In-context prompt, shown in Figure~(\ref{fig:icl_p}).
The Instruct prompt consists of a concise statement of the task goal, followed by direct input of all conversation histories into the model, serving as a test for the model's zero-shot capability. 
On the other hand, the In-context prompt involves presenting 4 examples prior to the samples to activate the model's imitation ability.

\end{document}